%% file: main.tex
\documentclass[conference]{IEEEtran}
\IEEEoverridecommandlockouts
\usepackage{cite}
\usepackage{amsmath,amssymb,amsfonts}
\usepackage{algorithmic}
\usepackage{graphicx}
\usepackage{textcomp}
\usepackage{xcolor}
\usepackage{comment}
\usepackage{acronyms}
\usepackage{misc}
\usepackage{balance}
\usepackage[export]{adjustbox}

\usepackage[caption=false]{subfig}

\def\BibTeX{{\rm B\kern-.05em{\sc i\kern-.025em b}\kern-.08em
    T\kern-.1667em\lower.7ex\hbox{E}\kern-.125emX}}

\begin{document}

\title{Marine Vehicles Localization Using Grid Cells for Path Integration
}

\author{Ignacio Carlucho\textsuperscript{a}, Manuel F. Bailey\textsuperscript{a}, Mariano De Paula\textsuperscript{b}, Corina Barbalata\textsuperscript{a} \\ 

icarlucho@lsu.edu, mbail37@lsu.edu,  mariano.depaula@fio.unicen.edu.ar, cbarbalata@lsu.edu \\

\textsuperscript{a } \textit{Department of Mechanical Engineering,} \textit{Louisiana State University}, Baton Rouge, USA \\

\textsuperscript{b }INTELYMEC Group, Centro de Investigaciones en F\'isica e Ingenier\'ia del Centro \\ CIFICEN  \textendash \space  UNICEN \textendash \space CICpBA \textendash \space  CONICET, 7400 Olavarr\'ia, Argentina \\

}

\maketitle
\thispagestyle{plain}
\pagestyle{plain}

\begin{abstract}

\acp{AUV} are platforms used for research and exploration of marine environments. However, these types of vehicles face many challenges that hinder their widespread use in the industry. One of the main limitations is obtaining accurate position estimation, due to the lack of GPS signal underwater.
This estimation is usually done with Kalman filters. 
However, new developments in the neuroscience field have shed light on the mechanisms by which mammals are able to obtain a reliable estimation of their current position based on external and internal motion cues. 
A new type of neuron, called Grid cells, has been shown to be part of path integration system in the brain.  
In this article, we show how grid cells can be used for obtaining a position estimation of underwater vehicles. The model of grid cells used requires only the linear velocities together with heading orientation and provides a reliable estimation of the vehicle's position. We provide simulation results for an \ac{AUV} which show the feasibility of our proposed methodology. 

\end{abstract}


\section{Introduction}

In recent years, the interest in exploring our oceans has increased. In this context, \acp{AUV} appear as an appealing option for performing underwater surveys autonomously.  
However, in underwater environments, since \ac{GPS} is unavailable, an accurate localization of the marine vehicle is hard to be obtain. Usually the location of the robot is estimated using information from on-board sensors such as the \ac{DVL} and \ac{IMU} \cite{barbalata2015adaptive},  but for long-term surveys error accumulates over time, causing high uncertainty in the vehicle navigation.
Another possibility is to get a reference measurement from vessels or buoys in the surface \cite{austin2000paradigm}. However, this technique can suffer from communication delays and requires additional infrastructure which increases the cost of the operation. Alternatively, biologically inspired methodologies such as the ones presented in \cite{Bioinspired2016} can provide an alternative for updating the robot localization based on internal motion cues.

\begin{figure}
 \centering
 \includegraphics[width=\linewidth]{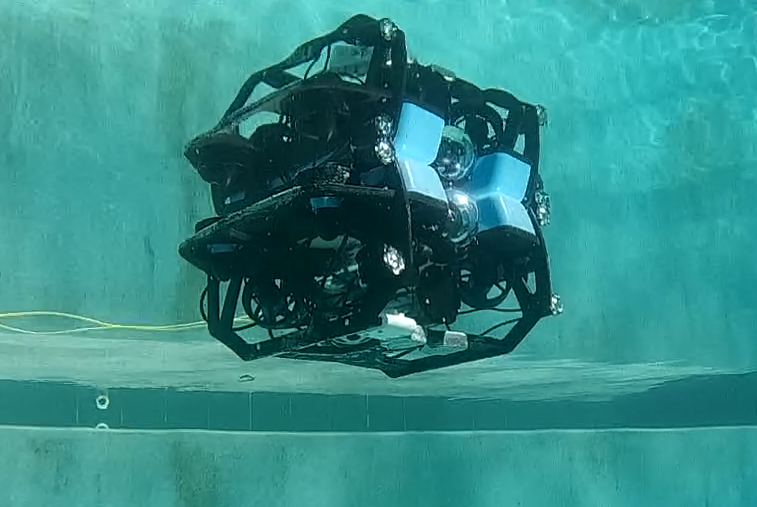}
 \caption{A modified BlueROV2 used for experiments}
 \label{fig:dory}
\end{figure}

Recent discoveries in the field of neuroscience have shed light on the process by which mammals are able to locate themselves in complex environments. 
It appears that, grid cells, a type of neuron found in the endorthinal cortex, are part of a general path integration system, that allows for self-localization \cite{Hafting2005MicrostructureOA}.  
These grid cells have a hexagonal pattern that strongly correlates with the animal position in space \cite{Zilli2012}. Evidence shows that, to navigate the environment, mammals use a combination of allothetic, such as external landmarks, and idiothetic cues, meaning internal information regarding the self motion \cite{POULTER2018R1023}. Furthermore, studies have shown that grid cells have a head direction preference, and their integration capability continues without the presence of visual cues, i.e., in complete darkness \cite{jacob:hal-02025122}.  

The concept of grid cells, used in neuroscience, can represent an enticing solution for the self-localization of robots in an unknown environment. 
In the literature, a limited amount of research has tested the applicability of this grid cell concept for robotic applications. 
In \cite{PioneerGrid} grid cells are used to navigate in an environment using a mobile robot, the Pioneer 3at, with the aid of visual cues. Their development is based on the RATSlam algorithm \cite{RATSlam}, which utilizes a simplified grid cell model. In \cite{Banino2018VectorbasedNU} it was shown that training an artificial agent using recurrent neural networks for solving navigation problems leads to the emergence of a grid-like representation in the network.
These initial works show promising results, and could potentially aid in solving the problem of inaccuracies in position measurements in underwater robotics. Nevertheless, existing algorithms and methodologies need to be adapted for the specific case of underwater robots.

The main contribution of this work is the analysis of grid cell 
as a localization algorithm for \ac{AUV} based on  
concept used in neuroscience. 
The central part of the localization algorithm is the use of a modified neural network that models the grid-like structure found in the endorthinal cortex. The structure of the network was originally presented in \cite{GUANELLA2007}. 
The network topology can be characterized by a twisted torus, in which the activity of rate-coded neurons are shifted by asymmetric synaptic connections. The inputs to the network are based on idiothetic information coming from the on-board sensors, i.e., velocity and heading. This information is used in the network to obtain an updated estimation of the current position, without the need of visual cues. The integration mechanism uses the changes in the state of the network as a way of updating the current position estimate of the \ac{AUV}, similarly as was done in \cite{PioneerGrid}. Additionally, we demonstrate the proposed approach to estimate the position of the \ac{AUV} shown in \figurename~\ref{fig:dory}. The obtained results show the feasibility of our proposed method. Furthermore, we include a discussion of the proposed methodology, considering future developments in the context of underwater vehicles.

The rest of the article is organized as follows. Section \ref{sec:related_work} presents related works. Section \ref{sec:grid} presents the methodology of the work, while Section \ref{sec:results} presents the obtained results. Finally, Section \ref{sec:discussion} presents a discussion, while Section \ref{sec:conclusions} presents the conclusions of this work.

\input{related}

\input{metodology}

\begin{figure*}[!ht]
    \centering
    \subfloat[Estimated position of the \ac{AUV} using grid cells versus the \ac{EKF} estimation]{%
      \includegraphics[height=6cm,valign=c]{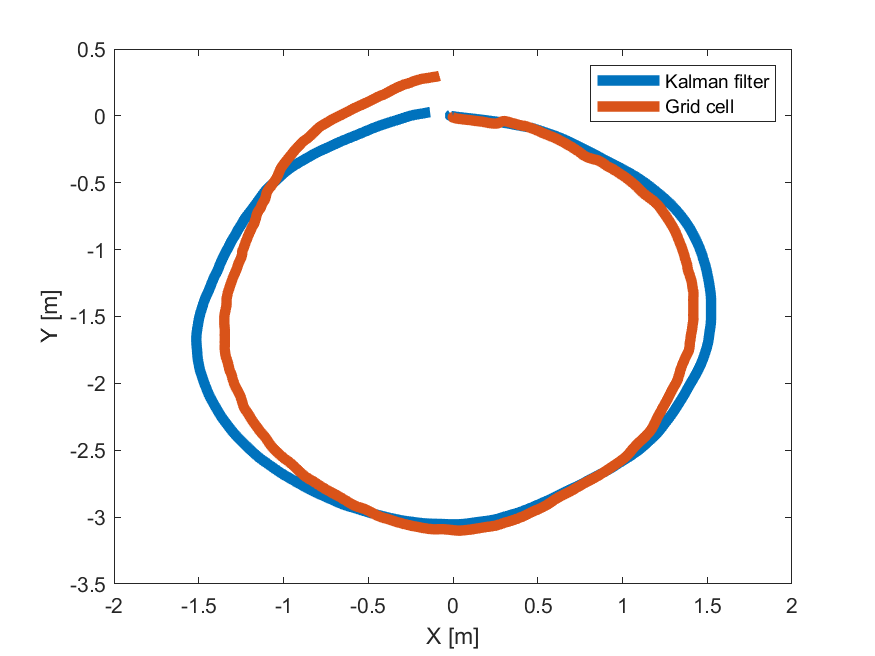}%
      \label{fig:position}%
    }
    \subfloat[Activity of the grid cell network over time]{%
      \includegraphics[height=5cm,valign=c]{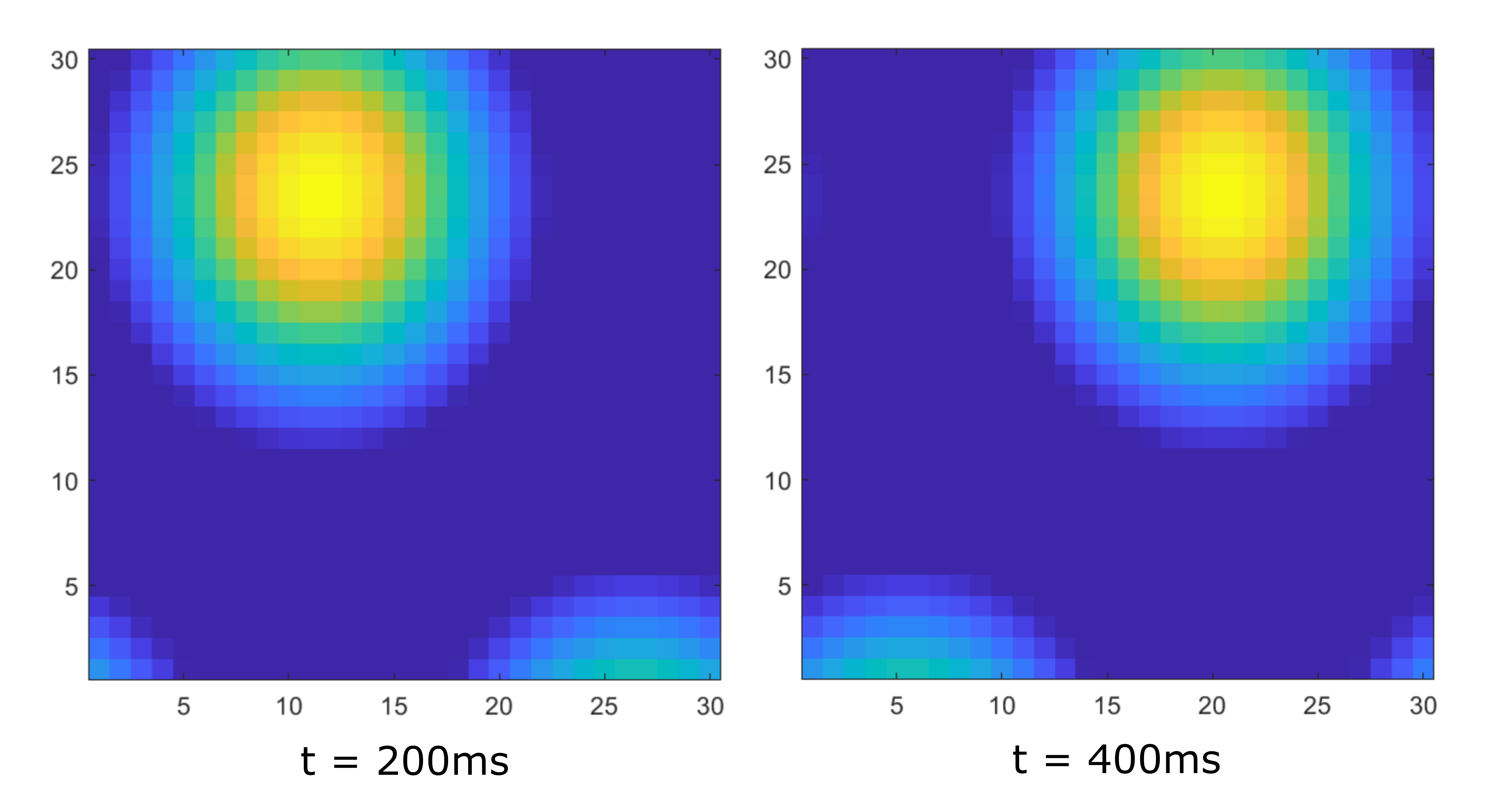}%
      \label{fig:grid}%
    } \hfil
    \caption{Results obtained using grid cells to estimate the position of an \ac{AUV} and comparison against an \ac{EKF}}
    \label{fig:circular_test}
\end{figure*}  
\input{results}

\input{discussion}

\section{Conclusions}
\label{sec:conclusions}

In this work, we present a grid cell model for path integration of underwater vehicles. The grid cell model utilizes a twisted torus topology, which directly takes as input the velocity of the vehicle. The output of the cells are shifted based on the velocity input, which allows to obtain an accurate position estimation. 

We tested the proposed approach in an underwater vehicle. The grid cell network takes as input directly the sensor readings, while at the same times it providing an accurate position estimation. The proposed scheme was tested under two scenarios, with and without disturbance. The obtained position estimation in both scenarios shows the capability of our proposed approach. 
In future works, we consider extending this methodology, combining the grid cell approach together with Kalman filter techniques to enhance the accuracy of pose estimation.

\section*{Acknowledgement}
We would like to thank Sean Katagiri from Heriot-Watt University, United Kingdom for helping us perform the experiments presented in this article. 
This work was supported by the EU H2020 Programme under the IRNOS Transnational Access of the EUMarineRobots project (grant ID 731103), and NSF Award $\#$2122068 RI: Small: Computational Imaging for Underwater Exploration.

\balance

\bibliographystyle{unsrt}
{\footnotesize \bibliography{references}}

\end{document}

%% file: related.tex
\section{Related Works}
\label{sec:related_work}



The ability for a robot to determine its position relative to an environment is critical in its ability to navigate an environment. The development of simultaneous localization and mapping, or SLAM \cite{Westman2018}, which, allows for a robot to simultaneously create a map of its environment and determine its pose relative to this map, changed the landscape of robotic localization. Early solutions to the SLAM problem consisted largely of probabilistic models. One of the first and most popular probabilistic models used in a SLAM implementation was the Kalman Filter \cite{kalman_new_1960}. The standard formulation of SLAM is found in \cite{Lu1997GloballyCR}. Recent improvement in computer technology has allowed for research on SLAM to focus on more robust methods, as the real world application of SLAM technology expands into a variety of real-world applications \cite{Cadena2016PastPA}. One promising area of SLAM research is the utilization of neural networks to obtain a reliable position estimation. In \cite{WangDeepVO} \ac{DRNN} are used for pose estimation by directly taking the inputs of a monocular camera. 

The theory that mammals store maps of the environment to assist their navigational abilities is first proposed in \cite{tolman_cognitive_1948}. The existence of spatial cells, namely place cells, in the hippocampus, and the idea that the hippocampus region of the brain functions as a cognitive map, was first introduced in \cite{okeefe_hippocampus_1971}. The function of grid cells as a spatial cell used for localization in the hippocampus was defined in \cite{Hafting2005MicrostructureOA}. While place cells and head direction cells create a good representation of current position, grid cells are useful for path integration mechanisms. It can be difficult to study the function of the cells directly from live animals, hence simulations and models have been developed in an attempt to refine our understanding of these cells. A multitude of models have been developed to explore the functionality of grid cells in the hippocampus, detailed in \cite{Zilli2012}. These models are broken down into two general categories. In the first, referred to as \ac{CANN} models, the grid cell activity of the model arises from a neural network. In the second, referred to as interference models, grid cell activity arises from individual cells, which store velocity information as a frequency. A number of models incorporate traits from both categories. In \cite{GUANELLA2007} the authors proposed a \ac{CANN} model of grid cell’s path integration function using an artificial neural network, and used a Gaussian weight function to describe the attractor dynamics of the system. 


The idea of using a model of the hippocampus for robotic navigation was first proposed in \cite{Burgess1997}. In \cite{Yu2019}, a complex model of the brain’s localization mechanism, which included grid, place, and head-direction cells, was implemented to solve the SLAM problem. It was an effective localization mechanism, but was not practical to replace exiting methods. As mentioned previously, the implementation of grid cell models in existing literature is explored in detail in \cite{Zilli2012}. The majority of models in literature that attempt to apply the function of spatial cells to robotic localization use \ac{CANN} models. A Bayesian attractor network model of grid cells and head-direction cells was proposed in \cite{ZENG202021}. This model which was shown to be a plausible localization method. A common drawback of models that use exclusively grid cell is that inaccuracies in the path-integration can compound over time without the recalibration provided by the other spatial cells. In \cite{Burak2009} the authors attempted to address this problem by proposing a path-integration method that does not accumulate significant error over time. However, the proposed model was dependent on very controlled operating conditions. In \cite{Milford2010} the authors proposed a method where the uncertainty can be effectively addressed by maintaining multiple probabilistic estimates of a robot’s pose.


%% file: metodology.tex
\section{Grid cells for navigation}
\label{sec:grid}

In this section we present the core aspects of our proposed methodology for position estimation. We start by providing an overview of continuous attractor networks, followed by a detailed description of the grid cell model utilized for path integration.

\subsection{Continuous Attractor Networks}

\acp{CANN} are a special class of neural networks, 
whose main difference 
is that the information in the network is encoded in firing patterns that correspond to stable states, also known as attractor states, of the network. 
An interesting property of these networks, that makes them suitable for modeling these type of biological systems, is that the hill of activity persists even in the absence of stimuli. 

The network structure is typically formed by a series of neurons and weighted connections between them. Some implementations include excitatory and inhibitory connections between nodes, looking to obtain local cooperation and distal inhibition \cite{kornienko_non-rhythmic_2018}. This causes neurons that are closer together to become mutually excited, increasing their activity, while the activity of distant neurons gets inhibited. This behavior causes bumps in activity that can be modified by the input to the neuron \cite{zeidman_neural_2008}. The shape of the peak can be modified by the weights of the neurons, while the localization of the activity in the network is guided by external inputs \cite{kornienko_non-rhythmic_2018}.
Particularly, \acp{CANN} have been used for modeling grid cells \cite{Burak2009}, head direction cells, \cite{stringer_self-organizing_2002}, and place cells \cite{conklin_controlled_2005}.

\subsection{Grid cell as continuous attractor networks}

As previously mentioned, \acp{CANN} have been used previously to represent grid cell model \cite{Shipston-Sharman2016}.
Particularly, the grid cell model utilized in this paper follows the developments presented in \cite{GUANELLA2007}. 
The model of the grid cell utilizes a number of $N$ neurons organized in a rectangular matrix, such that $N = N_x \times N_y$. This matrix represents the repetitive structure of grid cells subfields.

Additionally,  grid cells are divided into two different populations. The first is formed by the synapses that form the attractor dynamics on the grid cell. The synapses of this group connect cell $i$ and cell $j$, with $i,j \in \{1,2, ..., N \}$.
These connections between cells are modeled using a Gaussian weighted function. Connections between cell $i$ and cell $j$ are represented by $\omega_{ij}$. The individual weights are defined as: 
\begin{equation}
    \omega_{ij} = I \exp \left( - \frac{||\textbf{c}_i - \textbf{c}_j ||^2_{tri}}{\sigma^2}\right) - T
\end{equation}

\noindent where $I$ is an activation intensity parameter, $T$ is a shift parameter that affects the inhibitory and excitatory connections, $\sigma$ regulates the size of the Gaussian, $\textbf{c}_l$ defines the position of the cell $l$ on the grid, and the norm $|| \cdot ||$ defines the tessellation of the network \cite{GUANELLA2007}. Particularly, the position of the cell $\textbf{c}_l$ is defined as $\textbf{c}_l = (c_l^x, c_l^y)$, with $c_l^x = (l_x -0.5)/N_x$ and  $c_l^y = \frac{\sqrt{3}}{2}(l_y -0.5)/N_y$, 
where $l_x$ and $l_y$ are the column and row of cell $l$.
The second population is based on an external neuron $N+1$, to which all cells connect to, and works as a stabilization mechanism. This external cell is used to compute the mean activity of the grid cells group. The synapses of this group have a constant weight which is set equal to 1.

The dynamics of the grid cells neurons utilized in this work are governed by the following equation: 
\begin{equation}
  A_j(t+1) = f\left( (1-\tau)B_j(t+1) + \tau \left( \frac{B_j(t+1)}{\sum_{i=1}^N A_i(t)}\right)\right)
\end{equation}
\noindent where $A_i$ is the activity level of cell $i$, with $i \in \{1,2,...,N\}$, $\tau$ is a parameter that determines the stabilization strength, and $f$ is a simple rectification non linearity function, such that: $f(x) = x$ for $x>0$ and is $0$ otherwise \cite{GUANELLA2007}.  $\sum_{i=1}^N A_i(t)$ is computed locally using the activity of the external cell $N+1$. 
Additionally, the activity of the cells depends also on the transfer function $B_i$ defined as:
\begin{equation}
  B_j(t+1) = \sum_{i=1}^N A_i(t)w_{ij}(t)
\end{equation}
\noindent where $w_{ij}(t)$ are the weight connections from cell $i$ to cell $j$. The network is initialized with a random activity, uniformly distributed between $0$ and $1/\sqrt{N}$.

\subsection{Position estimation}

The previous description covers the fundamentals parts of the network. In this section we present how the velocity information is sent to the network and subsequently integrated to obtain a position estimation. %
The network takes as input a modulated version of the planar velocity. We define the velocity vector as: 

\begin{equation}
     \nu(t) = [v_x(t), v_y(t)]^T
\end{equation}

\noindent where $v_x(t)$ and $v_y(t)$ are the velocities in the $x$ and $y$ direction at time $t$. The input to the network is then modulated by a gain parameter and a rotation matrix such that:
\begin{equation}
    \nu^R(t) = \alpha \textbf{R}_{\beta} \nu(t)
\end{equation}

\noindent where $\textbf{R}_{\beta}$ is a rotation matrix that depends on the bias angle $\beta$ and the gain $\alpha \in \mathbb{R}^+$. It is important to note that the vector $\nu$ does not carry any information about the current position of the vehicle. When no velocity input is introduced, i.e. $\nu = [0,0]$, the grid remains stable. However, when the robot moves, this affects the activity of the network, shifting the bumps in activity. The effect of the velocity in the network is introduced in the synaptic weight equation, thus yielding: 

\begin{equation}
    \omega_{ij} = I \exp \left( - \frac{||\textbf{c}_i - \textbf{c}_j  + \nu^R(t) ||^2_{tri}}{\sigma^2}\right) - T
\end{equation}

It is possible to utilize the activity of the cell to obtain an updated estimation of the position. Considering at $t_0$ the position $x(t_0)=[0,0]$, the position is updated as: 
\begin{equation}
    x(t+1) = x(t) +\gamma \Delta A
\end{equation}
\noindent where  $\Delta A$ is the change in the activity bump, 
and $\gamma$ is a parameter that regulates the grid spacing.

%% file: results.tex
\section{Results}
\label{sec:results}

In this section we present the results obtained for the x-y position estimation of a BlueROV2 Heavy \cite{blue_robotics_2020} \ac{ROV} using the proposed approach. 
This portable \ac{ROV} is capable of operating at depths of up to 100m. The vehicle comes equipped with an \ac{IMU} sensor, which allows to estimate the current orientation of the robot. We modified the vehicle by including a \ac{DVL} sensor, the DVL A50 from water linked \cite{water_linked_dvl}, which allows to sense the linear velocities of the vehicle in the $x,\ y$ and $z$ axes. Additionally, we include a topside computer that uses \ac{ROS}, and that is able to read the sensor data and to send commands to the thrusters of the \ac{AUV}. This allow to deploy different control and navigation algorithms on the BlueROV2 using \ac{ROS}.  

The grid cell network was implemented in Matlab. We utilized similar hyperparameters as those presented in \cite{GUANELLA2007}. The full set of parameters utilized is presented in Table \ref{Tab:hyper}. The network is  directly fed with the data obtained from the \ac{DVL} and \ac{IMU}, namely, the body frame velocities in the $x$ and $y$ direction, rotated with respect to the current heading direction ($\psi$).

We compare the obtained behaviour with the position estimation when an \ac{EKF} is used based on the \ac{DVL} and \ac{IMU} measurements. We utilized the Robot Localization Package from \ac{ROS} to implement the filter. This package fuses the sensor data and provides a direct implementation of the \ac{EKF}, to obtain a position and velocity estimation.

\begin{table}[h]
    \caption{List of hyperparameters}
    \centering
     \begin{tabular}{c|c|c} 
     \hline
     Parameter & Value & Unit \\ 
     \hline\hline
     $N$ & 900 & [neuron] \\ 
     \hline
     $N_x$ & 30 & [neuron] \\
     \hline
     $N_y$ & 30 & [neuron] \\
     \hline
     $\tau$ & 0.95 & [no unit] \\
     \hline
     $\alpha$ & 1 & [no unit] \\ 
     \hline
     $\beta$ & 0 & [no unit] \\ 
     \hline
    \end{tabular}
    \label{Tab:hyper}
\end{table}

\subsection{Results without disturbance}

We first investigate the case when there are no external disturbances, the \ac{AUV} navigating in a tank with calm waters. The experiments were performed in a tank of 6 by 4 meters. We utilized the control and navigation system on board of the vehicle to perform different trajectories in the tank. The data from the sensors was recorded using ROS. The data was then processed offline and utilized as previously described. 

We present an experiment when performing a circular trajectory of 1.5 meters radius in the tank. \figurename~\ref{fig:circular_test} shows the robot's position estimation results when using grid cell approach and the \ac{EKF} methodology. 
It can be noted that while at first the position estimation given by the grid cell approach closely resembles that of the Kalman filter, the estimation starts to drift over time. Additionally, \figurename~\ref{fig:grid} shows the activity of the network during different moments of the test. The bright colors represent the activity bump of the network, that moves according to the velocity inputs.

\subsection{Results under disturbance}

\begin{figure}
 \centering
 \includegraphics[width=0.75\linewidth]{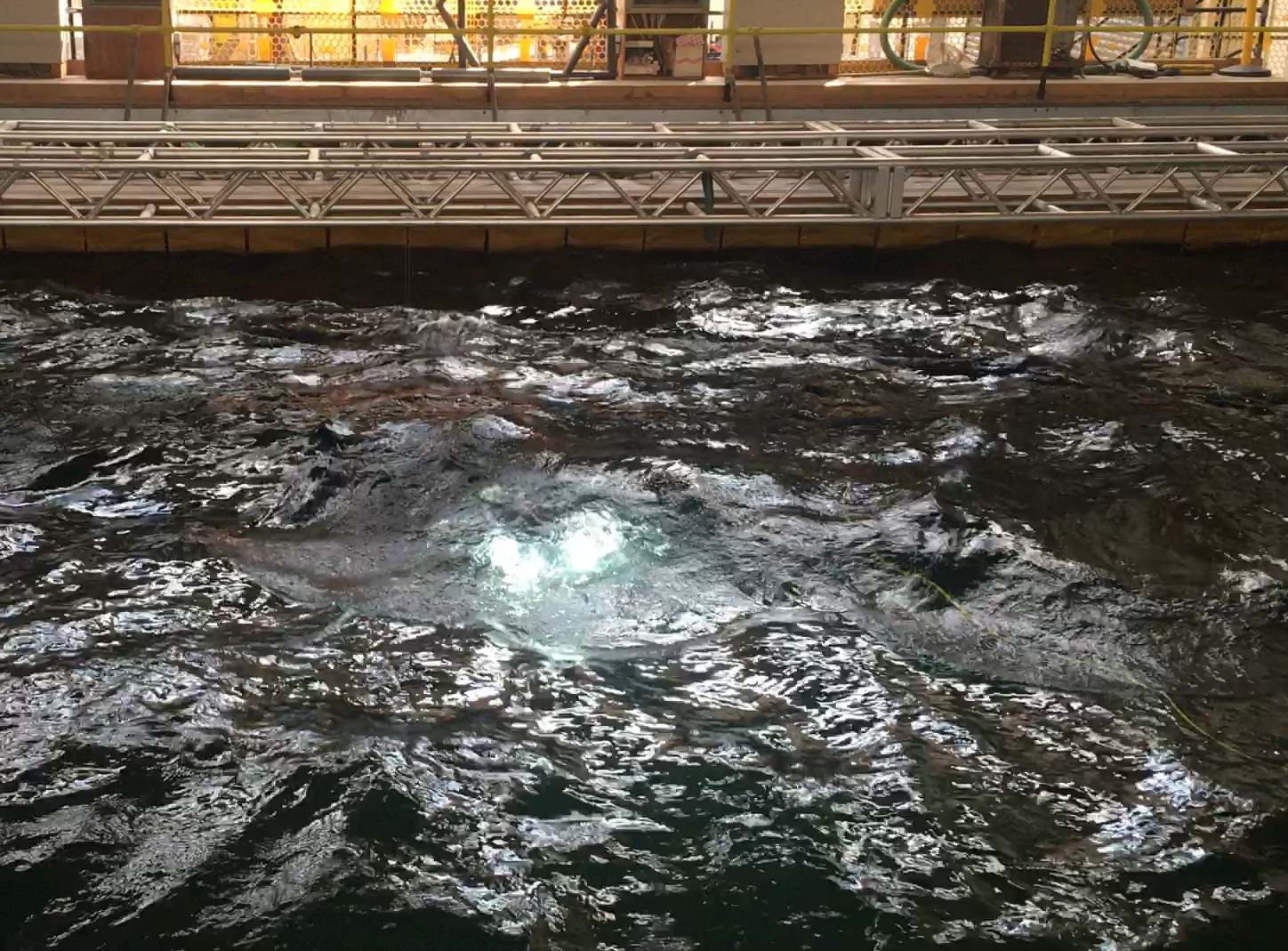}
 \caption{Wave tank testing platform utilized}
 \label{fig:wave_tank}
\end{figure}

In order to test the proposed navigation approach further, we performed a series of tests in a wave tank. This tank has the capability of producing waves at different frequencies, simulating various ocean currents, acting as disturbances on the vehicle. The tank and testing environment, located at Heriot-Watt University, can be seen in \figurename~\ref{fig:wave_tank}.  

For the this example, we require the vehicle to follow a series of waypoints leading to a rectangular trajectory. 

\begin{figure}
 \centering
 \includegraphics[width=\linewidth]{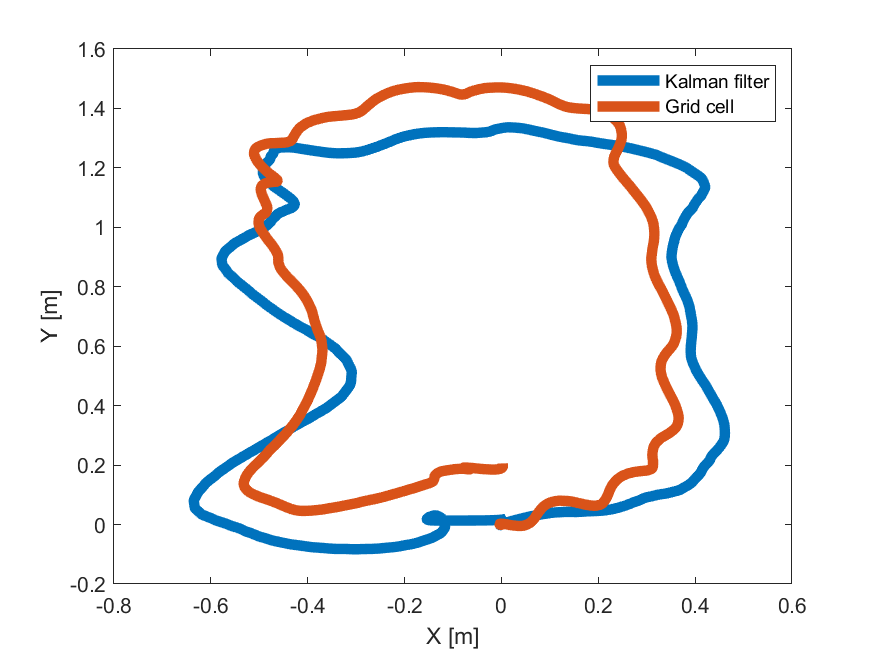}
 \caption{Position estimation of the \ac{AUV} using Grid cells versus the \ac{EKF} estimation, when perturbed by simulated ocean waves.}
 \label{fig:wave_results}
\end{figure}

The vehicle was asked to reach these waypoints while in the tank waves were generated at $1$hz. The obtained position estimation can be seen in \figurename~\ref{fig:wave_results}. We can observe how the estimation of the grid cell approach accumulates error more rapidly compared to the \ac{EKF} approach. From this test it can be seen that the proposed grid cell methodology is more sensitive to disturbances.

%% file: discussion.tex
\section{Discussion}
\label{sec:discussion}

The main advantage of this methodology is that it does not require a training stage, the grid cell behavior emerges from the population of neurons based on the defined structure of the network. 
Another benefit of this method is the low number of neurons utilized when compared with other neural networks based-approaches, such as deep neural networks \cite{DingDVLMALfunctions}. Furthermore, these types of neurons have the capability of learning and to update in a more efficient way, both energy and data wise \cite{SurveySNN}, but would require specific hardware \cite{SNNHardwareSurvey}, such as the Neurogrid \cite{Neurogrid} or TrueNorth \cite{Merolla2014AMS} to do so. 
Another advantage of this method is that it can be used in combination with other approaches. For example, the position estimation provided by the grid cells can be utilized together in combination with a Kalman filter, to obtain a more reliable position estimation. 
Nonetheless, while based on the grid cell architecture found in the medial enthorhinal cortex, the model 
used here constitutes an approximation which has limited biologically plausibility \cite{Zilli2012}. This is mostly due to the way the weights of the network are updated, which causes neurons in the grid to not have a fixed preferred direction.
However, other methods could be used to form a representation of grid cells \cite{ZENG202021}, utilizing a model less biologically plausible, such as Bayesian functions, but lowering the computational requirements for applications in commercially available hardware.

The proposed method was focused on idiothetic cues. However, allothetic cues may be used together with the implementation of place cells to increase the method's accuracy \cite{HarootonianPILandmark}. Typically, utilizing visual landmarks would be the most effective incorporation of allothetic cues \cite{Yu2019}. 
For underwater robots, in applications where the robot is required to navigate in the close proximity of the seabed, such as high resolution mapping or coral surveying, optical images could provide direct allothetic information allowing for the re-calibration of the position estimation. 
However, when navigating in the water column with limited visibility, external visual landmarks are not prominent, therefore other allothetic cues, such as acoustic signals, could be used to update the current estimated position.  
Other types of information, such as ocean currents or micro-particles in the water column, could provide allothetic information aiding the agent in generating a map of the environment, although this would require custom sensing capabilities.

Lastly, this work utilizes a planar representation of the space. This correlates with the way land mammals, such as mice, encode their position representation. However, studies have shown that this representation is different in fish and bats \cite{BurtdePerera2016}. For example, in mice, the cognitive map is highly resolved in the 2D plane when compared with the vertical direction. In contrast, in bats the spatial map shows a more spherical characteristic, showing that their representation of space is equally resolved in the 3D plane \cite{Yartsev367}. Similarly, studies show that the same type of representation happens in fish \cite{BurtdePerera2016}. A spatial representation of this sort would be helpful for an \ac{AUV} that is able to move in the 3D plane, however more studies are required to achieve this.